\documentclass[letterpaper, 10 pt, conference]{ieeeconf}  

\IEEEoverridecommandlockouts                              

\usepackage{graphics} 
\usepackage{epsfig} 
\usepackage{times} 
\usepackage{amsmath} 
\usepackage{amssymb}  
\usepackage{easyReview}

\usepackage[normalem]{ulem}

\usepackage{algorithm}
\usepackage{graphicx}
\usepackage{algorithmic}
\usepackage{times,subfigure}
\usepackage{caption}
\usepackage{subcaption}
\usepackage{adjustbox} 

\title{\LARGE \bf Bidirectional Progressive Neural Networks with Episodic Return Progress for Emergent Task Sequencing and Robotic Skill Transfer

}

\author{Suzan Ece Ada$^{1}$, Hanne Say$^{2}$, Emre Ugur$^{3}$, Erhan Oztop$^{4}$
\thanks{$^{1}$Suzan Ece Ada is with the Department of Computer Engineering, Bogazici University, Istanbul, Turkey
        {\tt\small ece.ada@bogazici.edu.tr}}%
\thanks{$^{2}$Hanne Say is with Department of Computer Science, Ozyegin University, Istanbul, Turkey
        {\tt\small hanne.say@ozu.edu.tr}}%
\thanks{$^{3}$Emre Ugur is with the Department of Computer Engineering, Bogazici University, Istanbul, Turkey
        {\tt\small emre.ugur@bogazici.edu.tr}}%
\thanks{$^{4}$Erhan Oztop is with SISReC, OTRI, Osaka University, Japan, and Department of Computer Science, Ozyegin University, Istanbul, Turkey
        {\tt\small erhan.oztop@otri.osaka-u.ac.jp}}%
}
\begin{document}

\maketitle
\thispagestyle{empty}
\pagestyle{empty}

\begin{abstract}
Human brain and behavior provide a rich venue that can inspire novel control and learning methods for robotics. In an attempt to exemplify such a development by inspiring how humans acquire knowledge and transfer skills among tasks, we introduce a novel multi-task reinforcement learning framework named Episodic Return Progress with Bidirectional Progressive Neural Networks  (ERP-BPNN). The proposed ERP-BPNN model  (1) learns in a human-like interleaved manner by  (2) autonomous task switching based on a novel intrinsic motivation signal and,  in contrast to existing methods, (3) allows bidirectional skill transfer among tasks.  ERP-BPNN is a general architecture applicable to several multi-task learning settings; in this paper, we present the details of its neural architecture and show its ability to enable effective learning and skill transfer among morphologically different robots in a reaching task. The developed Bidirectional Progressive Neural Network (BPNN) architecture enables bidirectional skill transfer without requiring incremental training and seamlessly integrates with online task arbitration. The task arbitration mechanism developed is based on soft Episodic Return progress (ERP), a novel intrinsic motivation (IM) signal. To evaluate our method, we use quantifiable robotics metrics such as `expected distance to goal'  and `path straightness'  in addition to the usual reward-based measure of episodic return common in reinforcement learning. With simulation experiments, we show that ERP-BPNN achieves faster cumulative convergence and improves performance in all metrics considered among morphologically different robots compared to the baselines. 
\end{abstract}

\section{Introduction}
Developing robots capable of autonomous and continual learning effectively requires the exploitation of acquired knowledge without human intervention, which could be described as the main goal of lifelong robot learning \cite{oztop_lifelong_2020,parisi_continual_2019,thrun1995lifelong}. A key feature of human learning is autonomous task switching and interleaved learning, which is not addressed in mainstream machine learning and robotics research \cite{say2023model}. In this study, we aim to help fill this gap by developing a multi-task reinforcement learning framework that can sustain and, importantly, benefit from interleaved task learning and allow bidirectional skill transfer among tasks. 

In humans, it is shown that interleaved learning yields improved recall of information and better memory retention in the long run rather than blocked learning \cite{lin2013interleaved, samani2021interleaved, park2023role}. Supporting this behavioral data, the human brain is endowed with mechanisms against task interference and forgetting, such as internal rehearsal of experiences and memory consolidation \cite{nadel2012memory, cepeda2008spacing}. To enable interleaved learning, the question of when and which task to engage during multi-task learning must be answered. Developmental learning offers some inspiration: during learning, an infant autonomously decides what to do or play without external directions dictating what task (s)he needs to engage in. This behavior of infants is usually associated with the notion of \emph{intrinsic motivation} (IM), which guides behavior through a putative internal reward system \cite{oudeyerautonomous}. IM can be based on curiosity, novelty, learning progress (LP), or even challenge; as such, it is also used in robotics as LP \cite{pmlr-v97-colas19a}, curiosity \cite{forestier2016curiosity}, and in machine learning as novelty \cite{gatsoulis2015intrinsically}, or surprise \cite{achiam2017surprise}. In the current study, we also adopt an IM approach and propose a novel learning progress (LP) signal for RL tasks that guide task switching. Overall, inspired from the above discussion, we aim to develop a multi-task reinforcement learning (RL) framework with autonomous task switching, which can benefit from interleaved task learning without suffering from task interference. We argue that such a learning system may benefit a wide range of robot learning scenarios ranging from human-like learning for a social robot to skill transfer applications among morphologically different robots. 

While significant progress has been made in deep reinforcement learning (RL) for robotics, most approaches \cite{varghese_surveymtl_2020} have focused on transferring skills between robots with identical action spaces \cite{zhiyuan_knowledge_transferNEURIPS2020,teh2017distral} and tasks with pixel-level state inputs \cite{Zhang_deeprl_successor_2016,yin_knowledgetransfer_2017,liu2016decoding}. Nevertheless, achieving multi-task reinforcement learning between robots with different physical structures provides insights for human-inspired reinforcement learning and acts as a critical driving force for machine learning research \cite{roy_machine_2021}. This is particularly significant due to the different state and action spaces, providing a unique perspective on knowledge generalization. In light of this, we choose learning and transferring skills among morphologically different robots as the target to address with the the developed framework. 

One of the key challenges in lifelong learning is catastrophic interference/forgetting, which needs to be taken into consideration if a robot is to learn continually. When novel instances to be learned diverge greatly from previously observed ones, new information may overwrite the already acquired knowledge by modifying the representations that are shared among multiple tasks, leading to catastrophic forgetting. To minimize the interference while learning a novel task, in the literature, several techniques have been proposed, such as restricting the probable update(s) on the network parameters, dynamic resource allocation, or rehearsing the old task samples while learning the new ones \cite{oztop_lifelong_2020}. In this study, similar to the Progressive Neural Networks (PNN) \cite{rusu_progressive_2016}, we utilize task-specific resources while learning to prevent possible task interference but also allow bidirectional inter-task connectivity to support positive skill transfer. 

\begin{figure}[!htbp]
\centering
        \subfigure[]{
		\includegraphics[width=0.3\columnwidth]{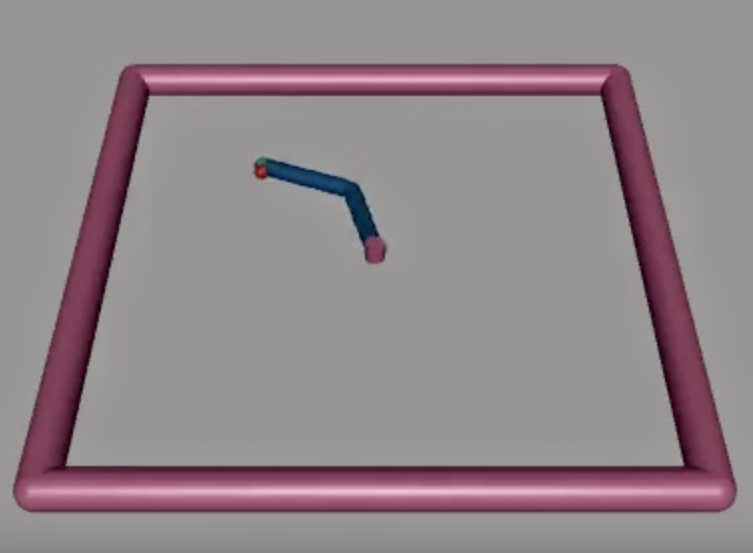}}
        \subfigure[]{
		\includegraphics[width=0.3\columnwidth]{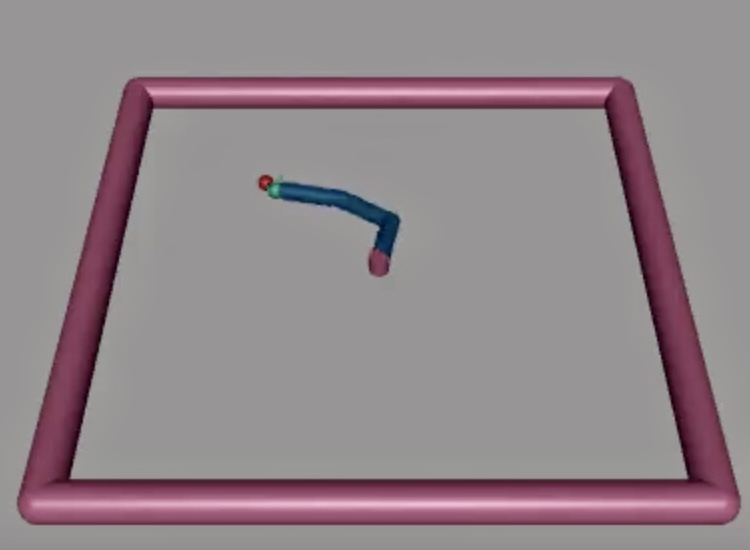}}
        \subfigure[]{
		\includegraphics[width=0.3\columnwidth]{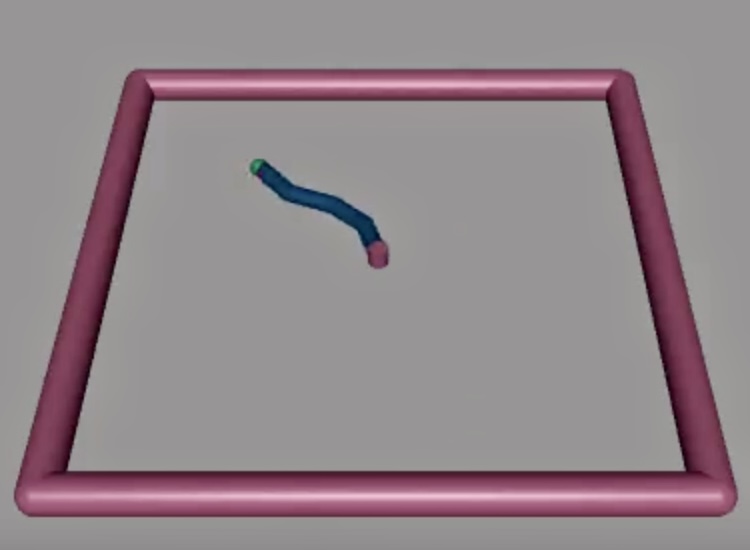}}
	\caption{(a) 2-DoF, (b) 3-DoF, (c) 4-DoF Reacher Robot Arm Environments}
	\label{fig:robotfigs}
\end{figure}

In sum, to enable human-like interleaved multi-task learning while avoiding task interference, we develop a multi-task reinforcement learning system composed of  (1) a novel architecture called BPNN that improves the PNN \cite{rusu_progressive_2016,rusu_simtorealpnn_2017} and  (2) a novel Intrinsic Motivation signal, Episodic Return Progress (ERP), for task-switching. Unlike the PNN architecture, which restricts skill transfer to the forward direction and requires learning previous tasks until convergence before transferring to the next task, our BPNN method enables bidirectional skill transfer during training. This means that skill transfer can happen in midway along multi-task learning among all tasks without the necessity of one task waiting for another to finish. The ERP signal evaluates task progress based on episodic return values, detecting the task that significantly contributes to enhancing overall performance across multiple tasks. The efficacy of the proposed multi-task learning framework is shown by its application to the learning of reaching skill by morphologically different robots, namely two degrees of freedom (2-DoF), 3-DoF, and 4-DoF manipulators (Figure \ref{fig:robotfigs}). The conducted systemic experiments show that synergistic multi-task learning is possible due to bidirectional inter-task skill transfer provided by the proposed BPNN architecture and the ERP-based task switching.

The rest of the paper is organized as follows. In Section \ref{sec:RELATED}, we present an overview of the related studies present in the literature. Then, we describe our method in detail in Section \ref{sec:METHOD}, providing metrics used to evaluate the performance of our proposed method. Section \ref{sec:EXPERIMENTS} details the experiments and presents the results for skill transfer between morphologically different reacher robots. Finally, we discuss the broader impact of our method and future research directions in Section \ref{sec:CONCLUSION}.

\section{RELATED WORK}
\label{sec:RELATED}
\paragraph{Human Learning}
Numerous examples demonstrate how neuroscience and artificial intelligence have paved the way for each other \cite{hassabis_neuroscience-inspired_2017,niv2009,khetarpal_towards_2022}. In this vein, the ability of humans to acquire multiple skills with ease throughout their lives may guide machine learning research in multi-task learning and lifelong learning fronts. 
Human learning, especially during infancy and childhood, is characterized by autonomous engagement in play, i.e., exploration, where no task is learned completely in one sitting. Besides being ecologically unreasonable to focus on learning a single task until mastery, interleaved learning may allow positive skill transfer among tasks from partial learning if adequate mechanisms are engaged. This notion is supported by the contextual-interference (CI) effect studies showing that practicing tasks in an interleaved regime often results in improved learning compared to practicing in a block order \cite{brady2008contextual}. Benefits of CI have been linked to increased brain activity during interleaved practice as opposed to repetitive practice \cite{cross2007neural, lin2011brain}. In addition, the CI effect not only improves information retention but also skill transfer between similar tasks
\cite{schorn2021interleaved}.

On the other hand, machine learning settings usually prefer blocked learning to interleaved learning. Multi-task learning settings generally assume that either a random task (or subset of tasks) is chosen for each training trial or that training proceeds to the next task after one task is mastered with a few exceptions \cite{sharma2017learning, jean2019adaptive}. In the case of continual learning, tasks are learned sequentially as each task arrives \cite{lesort_continual_2020}. This is in contrast with how humans learn. It is clearly seen that the beneficial impacts of interleaved learning have not been thoroughly examined within the machine learning literature thus far. Investigating these effects has the potential to enhance the alignment of artificial intelligence with the underlying mechanisms of the human brain.

\paragraph{Intrinsic Motivation}

Intrinsic motivation (IM) is a significant topic in infant cognitive development and learning, which refers to motivation originating from innate satisfaction instead of the extrinsic reward gained from the environment \cite{oudeyer2007,kasmarik2023guest}. IM has been adopted for enabling open-ended robot learning \cite{santucci2020intrinsically}, and improving self-supervised exploration \cite{sener2021exploration, Bugur}. Intrinsic rewards can arise from exploring novel states, satisfying an ingrained curiosity, or the rate of acquiring skills and knowledge in an environment. Inclusion of intrinsic rewards to the extrinsic rewards from the environment is one way of solving challenging exploration problems \cite{burda2018exploration} and discovering a diverse set of skills \cite{eysenbach2018diversity} in deep reinforcement learning.  In our approach, unlike the existing uses of IM in reinforcement learning, we propose to utilize episodic return progress (ERP) as a higher-level IM signal to dynamically switch tasks for learning in an online manner instead of modulating the reward signal guiding RL.  Consequently, the dynamic task switching carried out by the task selection mechanism leads to an emergent interleaved multi-task learning regime.

\paragraph{Curriculum Reinforcement Learning}
Curriculum learning methods focus on discovering a goal or task sequencing procedure that can lead to a faster convergence during training or improved performance compared to random sequencing \cite{bengio_curriculum_2009}. Most curriculum learning techniques require a priori domain knowledge of tasks to distinguish task levels to train from easier tasks to more difficult tasks \cite{narvekar_curriculum_2020,weinshall2018curriculum,Ugur2014}. For example, for predicting the output of a short Python code with Long Short Term Memory (LSTM) networks, the task levels can be identified by the number of nestings and the number of digits in the integers \cite{zaremba2014learning}. After manually identifying these task difficulty measures, it can be demonstrated that while training, a combination of a random curriculum and a naive curriculum where tasks are selected in ascending order of difficulty performs better than using only a random or a naive curriculum  \cite{zaremba2014learning}.
However, the difficulty of each task might not be readily available a priori in the robotics domain, and thus an automatic or emergent curriculum formation can be desirable. Deep Q-Networks (DQN) with prioritized experience replay \cite{schaul_prioritized_2016} assigns importance to transitions based on their associated temporal difference error, thereby selecting data for more frequent replay in single-task reinforcement learning. On the other hand, our method prioritizes which task network is allowed to learn in an online manner. Hence, it operates at a higher level than DQN with prioritized experience replay's prioritization scheme and creates emergent interleaved task-switching patterns on the fly. As such, the learning scheduling obtained is quite different from what can be obtained through a transition-level curriculum or a usual task curriculum where each task is learned to completion.

\paragraph{Multi-task Reinforcement Learning}

Multi-task learning involves sharing skills and knowledge between multiple tasks, where each task is identified as either a source or a target task during training. Typically, tasks share a part of the neural network model, and the model integrates a task conditioning parameter that defines the task during training. PathNet is a technique that uses a tournament selection genetic algorithm to evolve pathways of a neural network for multi-task, lifelong, and forward transfer learning \cite{fernando_pathnet_2017}. However, Pathnet is trained consecutively for reinforcement learning tasks, meaning that the source task needs to be trained until convergence before moving on to the target task. This procedure does not allow backward transfer of tasks. Similarly, in PNN \cite{rusu_progressive_2016}, training the source task until convergence is required to transfer skills to other tasks that are connected to the trained task. In contrast, ERP-BPNN supports bidirectional transfer and does not require convergence in one task to use previously learned representations in other tasks. Hence, ERP-BPNN fits better into the multi-task learning framework where all tasks are source and target tasks throughout training. This is essential because the BPNN architecture allows for integrating learning progress and bidirectional transfer. 

\section{METHOD}
\label{sec:METHOD}
We propose a novel multi-task reinforcement learning framework that integrates a bidirectional progressive neural network (BPNN) with a unique architecture and soft task-switching mechanism crafted for RL, inspired by our previous work \cite{say2023model}. The BPNN architecture consists of bidirectional lateral connections among the hidden layers of each fully connected task network to allow skill transfer. By allocating a separate network module for each task, the model avoids negative task transfer at the core level but allows potential positive transfer to take place due to the bidirectional lateral connections among networks. 

\begin{figure}[!htbp]
\centering
        \subfigure[]{
		\includegraphics[width=0.4\columnwidth]{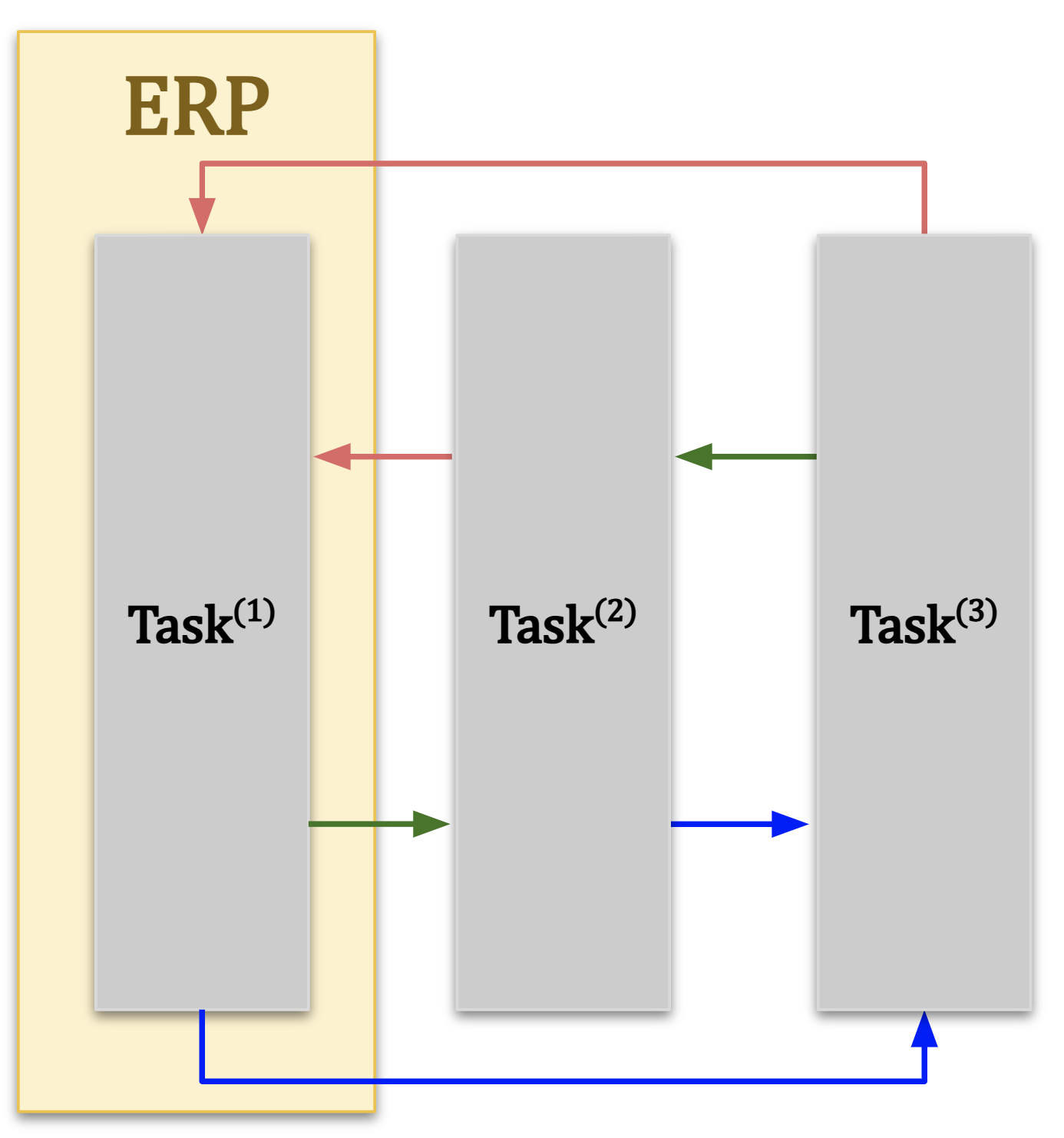}}
        \subfigure[]{
		\includegraphics[width=0.373\columnwidth]{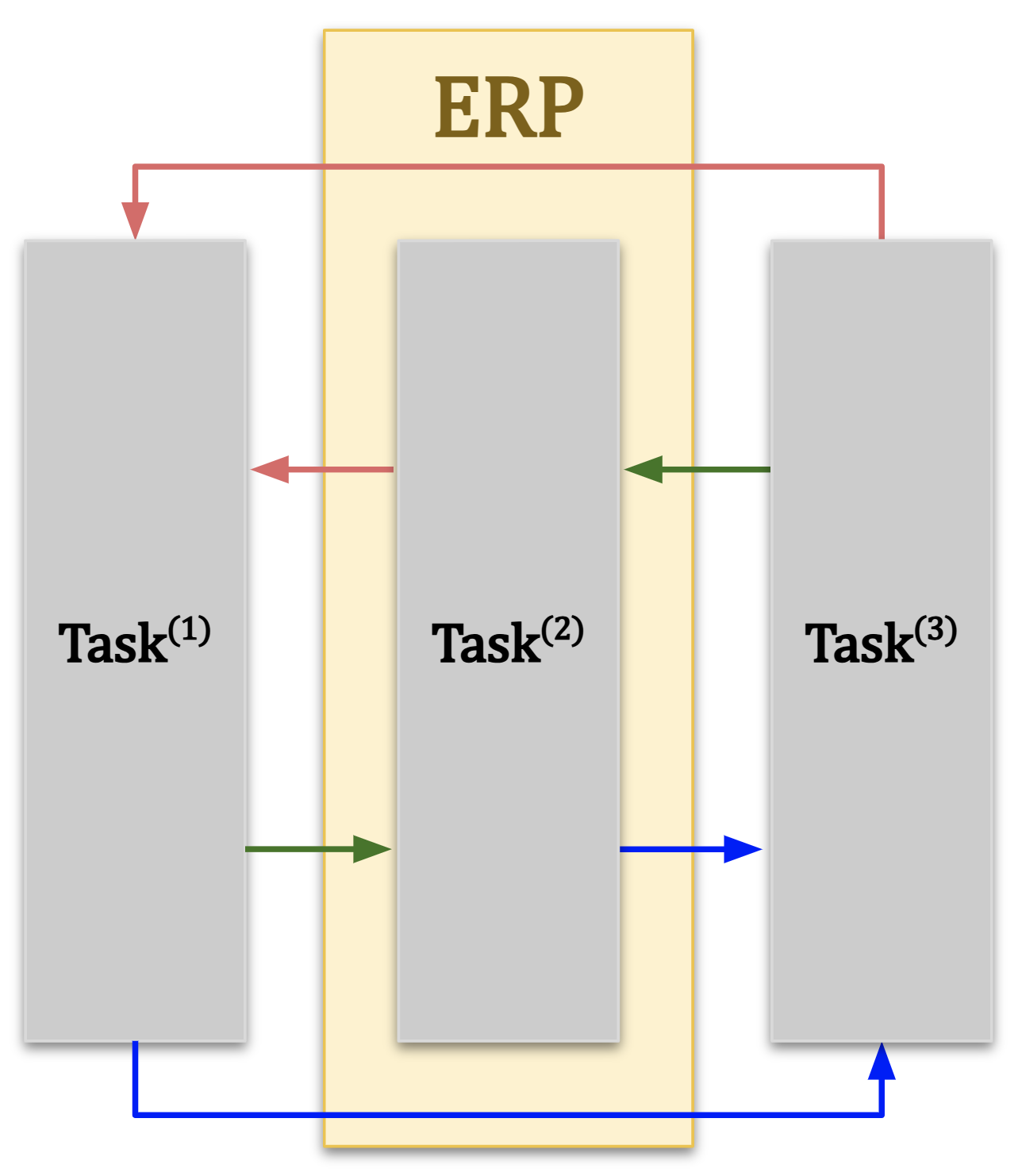}}
        \subfigure[]{
		\includegraphics[width=0.8\columnwidth]{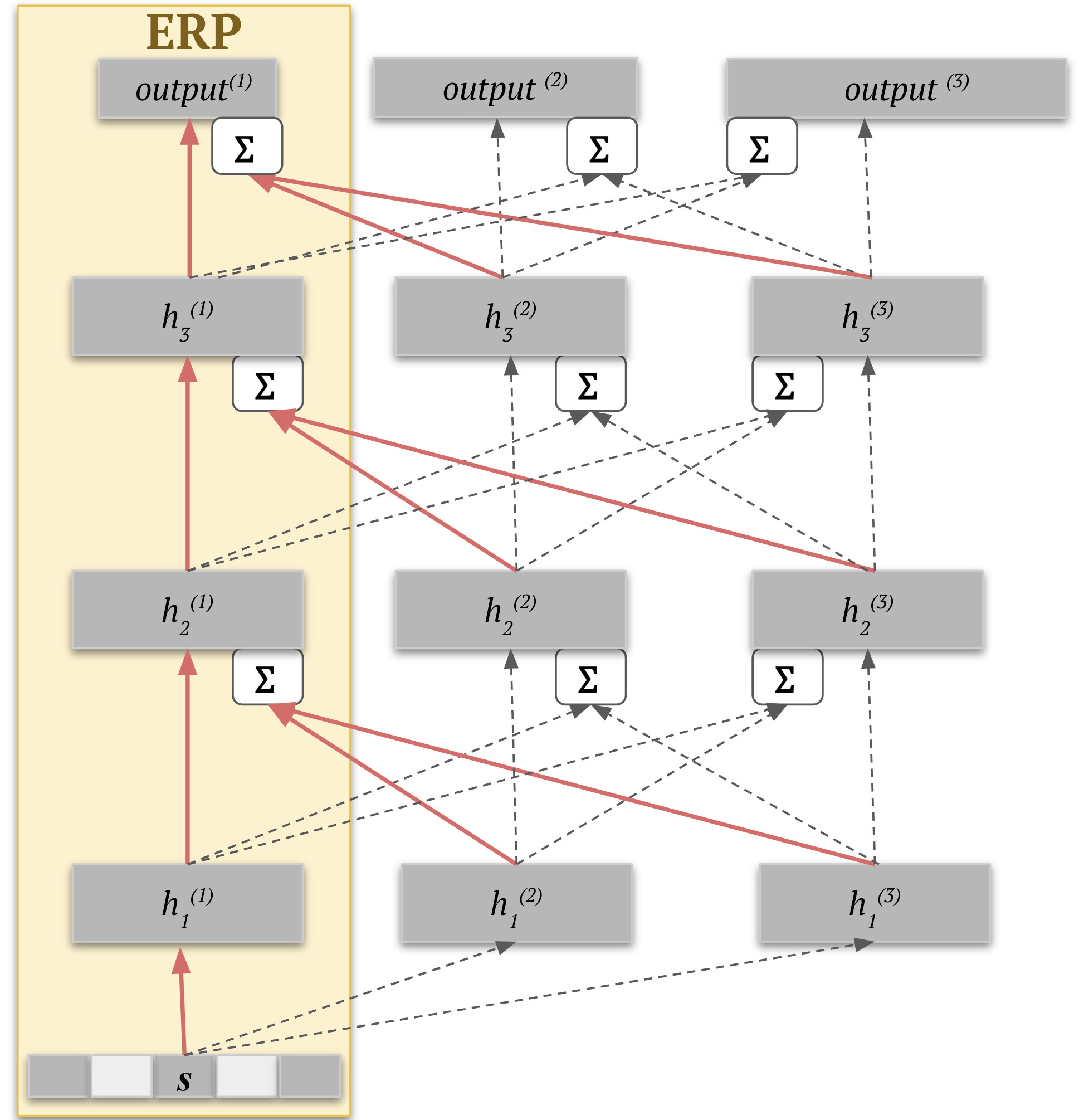}}
	\caption{A high-level (a,b) ERP-BPNN architecture demonstrates ERP selecting Task 1 (a) then Task 2 (b), showcasing bidirectional flow for skill transfer in a many-to-many fashion among three tasks. Graphical representation of ERP-BPNN framework with ERP task switching wherein (c) Task 1 is selected to learn. Weight updates are denoted by red (Task 1) arrows. Dashed gray arrows indicate no gradient flow during the learning update. In the current report, Task 1, 2, and 3 refer to RL tasks for 2-Dof, 3-DoF, and 4-DoF Reacher Robot arms as illustrated in Fig. \ref{fig:robotfigs} (a), (b), (c) respectively.}
\label{fig:erpbpnnarchitecture}
\end{figure}

\subsection{Bidirectional Progressive Neural Networks}

We initialize a fully connected neural network module $m \in M$ with task parameters $\theta_m$ for each task $\mathcal{T}_i$ with index $i$. As each task corresponds to a module, we will use ``network module" and ``task" terms interchangeably. Since each robot controls different numbers of joints, each network module has task-specific output layers with a different number of neurons $n^{m}_{\{l=L\}}$ where $l \in \{1..L\}$ refers to the layer index. 

During training, each network module receives the input of the task selected for training. In this way, the lateral activations can receive representations of other tasks for skill transfer. Subsequently, we compute the hidden activations $h_l^{(m)} \in \mathbb{R}^{n^{m}_l}$

\begin{equation}
\scriptstyle{
h_l^{(m)}=f\left(W_l^{(m)} h_{(l-1)}^{(m)}+b_l^{(m)}+\sum_{t\neq m}U_l^{(m:t)} h_{l-1}^{(t)}+b_l^{(m:t)}\right)}
\end{equation}

where, $U_l^{(m:t)}$, $b_l^{(m:t)}$ denote the weight matrix and the bias vector corresponding to the lateral connections of the ordered pair of modules $(m,t)$ from the previous layer $h_{l-1}^{(t)}$ of the module $t$ to the $l^{th}$ layer of module $m$. $W_l^{(m)} \in \mathbb{R}^{n^{m}_l \times n^{m}_{l-1}} $ and $b_l^{(m)}$ are the weight matrix and bias vector, respectively, for layer $l$ of module $m$, and $f$ is the element-wise activation function. In the BPNN architecture, linear layers are utilized to potentially transmit information from other tasks, represented by $\boldsymbol{\Sigma}$ in Fig. \ref{fig:erpbpnnarchitecture}(c), functioning as a module for summation. The linear transformations applied to the previous layers of other tasks allow the tuning of the incoming lateral signals. In particular, a negative transfer can be suppressed and a positive transfer can be enhanced by the tuning of lateral weights through gradient descent-based learning. In this work, we use the activation function $tanh(x)=\frac{e^x-e^{-x}}{e^x+e^{-x}}$ following the suggested hyperparameters for PPO \cite{schulman_proximal_2017} for continuous control. In the initial phase, the lateral connections of each module are frozen, and all tasks are individually trained to jumpstart task-specific learning for a predetermined number of $K_{init}$ training iterations. 
In the experiments reported in Section 4, we set $K_{init}=20$. In the subsequent learning steps, to avoid negative transfer between tasks, the parameters of the task modules that are not selected for training are frozen. This prevents the gradient flow into the networks associated with the remaining tasks. On the other hand, the parameters of and the lateral connections incoming to the task module selected for training are unfrozen to facilitate task-specific learning as well as the skill transfer from other tasks.
How the dynamic task selection takes place is described next.
\vspace{3mm}

\subsection{Task Switching by a Novel Intrinsic Motivation Signal: Average Soft Episodic Return Progress}

We propose a task-switching mechanism for multi-task reinforcement learning based on a novel  Intrinsic Motivation (IM) signal, namely Average Soft Episodic Return Progress (ERP), that captures the learning progress of an agent in the RL context. At each optimization iteration $k$, for each task network module $m$, we record the expected discounted cumulative reward or average episodic return denoted by $R_m(k)$. To compute $R_m(k)$, we first normalize the immediate rewards to ensure the exponential moving average of the rewards has a constant variance and clip them between (-10,10) to stabilize training \cite{huang2022cleanrl}. Then we compute the mean of the returns over $\boldsymbol{P}*\kappa$ trajectories where $\boldsymbol{P}=8$ is the number of parallel RL environments, and $\kappa=2$ is the number of episodes completed in each RL environment. Then, we take ERP for task $m$ at step $k$, $ERP_m(k)$, as the slope of the line that is fitted to $R_m(k-w+1), R_m(k-w+2), \hdots, R_m(k)$ using least squares, where $w$ is a predetermined window size ($w=5$ in the experiments reported in this paper). The least squares solution is given by 

\begin{equation}
\scriptstyle{
ERP_m(k)=\frac{w\left(\sum^{w-1}_{i=0} X_k\lbrack i\rbrack Y_m(k)\lbrack i\rbrack\right)-\sum^{w-1}_{i=0} X_k\lbrack i\rbrack \sum^{w-1}_{i=0} Y_m(k)\lbrack i\rbrack}{w\left(\sum^{w-1}_{i=0} \left(X_k\lbrack i\rbrack\right)^2\right)-\left(\sum^{w-1}_{i=0} X_k\lbrack i\rbrack \right)^2}}
\end{equation}
where $X_k=[k-w+1,k-w+2,.....,k]$ and $Y_m(k)=[R_m(k-w+1),R_m(k-w+2),.....,R_m(k)]$.

For bootstrapping ERP computation for each task, at the beginning of a multi-task learning session, each task is given an initial ERP bootstrapping run of $K_{init}>w$ iterations. Then, we compute the ERP for each module individually for the initial run and subsequently at each iteration to select the task that has made the highest recent progress. The objective of the ERP procedure is to select tasks dynamically by identifying the task with the highest recent progress so that the selected task can continue its rapid progress. Selecting the most efficient task for training is particularly important as it can facilitate skill transfer to the other tasks,  utilizing the bidirectional lateral connections of BPNN. For instance, the tasks that have made less progress can benefit from the steeper improvement of the selected task. Since ERP is computed for all tasks at each iteration, other tasks can benefit from the top-performing task and increase their chances of getting selected. The window size should be chosen to balance the reduction of noise and the tracking of recent updates. A large window size, $w$, can lead to selecting the initially top-performing task for an extended number of iterations. Correspondingly, a small window size can introduce noise during training. Thus, to monitor the recent changes in the progress of all tasks, we empirically selected the window size as five based on a grid search. Consecutive selection of the same task will eventually reach a plateau during training either due to the fact that the task is learned or a local minimum is encountered,  where learning in other tasks may help the plateaued task to jumpstart as the reduction in progress of the current task allows other tasks with more progress to be selected. This dynamic task selection regime can be considered analogous to the flow state theory \cite{csikszentmihalyi_1990}, which strives to maintain a balance between challenging and effortless tasks. 

\subsection{Multi-task Reinforcement Learning}
We initialize two separate BPNN architectures for critic and policy networks to integrate our method into the actor-critic reinforcement learning framework. In this manner, the policy network receives the state for the corresponding task as input and outputs the mean of a diagonal multivariate Gaussian distribution with a learned log standard deviation parameter independent of the state. Correspondingly, the critic network receives the state and learns the value function. Hence, only the parameters of the training task's actor-critic modules and their corresponding lateral connections are updated using Adam optimizer \cite{2015-kingma} during task learning. We use the tuned hyperparameters in \cite{rl-zoo3} for the reacher task as there are multiple extensions of PPO \cite{Engstrom2020Implementation} and compute advantage estimates $\hat{A}_t$ with \cite{pmlr-v48-mniha16}. This extended PPO loss $L^{\mathcal{T}_i}_{PPO}$ with a clipped value function loss and a value function entropy bonus can be formulated \cite{schulman_proximal_2017} as 

\begin{equation}
\scriptstyle{
L^{\mathcal{T}_i}_{PPO}=\mathbb{E}_t[L^{CLIP_{\mathcal{T}_i}}_t(\theta)-c_1L^{VF_{\mathcal{T}_i}}_t(\theta)+c_2S^{\mathcal{T}_i}(\theta(\pi)(s_t)]}
\label{eq:ppo}
\end{equation}

\noindent where $L^{CLIP_{\mathcal{T}_i}}_t(\theta)$ is the clipped PPO surrogate objective, $c_1L^{VF_{\mathcal{T}_i}}(\theta)$ is the clipped value loss with coefficient $c_1$, $c_2S^{\mathcal{T}_i}(\theta(\pi))(s_t)$ is the entropy bonus for the actor network with coefficient $c_2$. Although the actor and critic neural networks are separate, we can denote them collectively as $\theta$ as in previous works \cite{schulman_proximal_2017} for brevity. For instance, $\theta(\pi)$ and $\theta(\phi)$ refer to the actor-network and critic-networks parameters, respectively.

\begin{algorithm}
    \caption{ERP-BPNN}\label{erpbpnnalgorithm}
    \begin{algorithmic}[1]
    \REQUIRE $\mathcal{T}_i$: Task with index $i$, $\mathcal{T}$: set of tasks, $m_{\mathcal{T}_i}$: network module of $\mathcal{T}_i$ \\
    \STATE Constants: $\kappa=2$ (\#episodes), $P=8$ (\#parallel tasks), $K_{init}=20$ (\#jumpstart training iterations)
        \FORALL{$\mathcal{T}_i$}
        \STATE Unfreeze module $m_{\mathcal{T}_i}$
         \FOR{$k \in \{1,...,K_{init}\}$}
           \FOR{$p \in \{1,...,P\}$} 
                \STATE Sample $\kappa$ episodes by running policy $\pi_{\theta^{old}_{\mathcal{T}_i}}$
                \STATE  $\forall t,$ Compute Advantage estimates $\hat{A}_t$
            \ENDFOR
            \STATE Record average episodic return $R_{m_{\mathcal{T}_i}}(k)$
            \STATE Optimize $L^{\mathcal{T}_i}_{PPO}$ w.r.t. $\theta_{\mathcal{T}_i}$ by Equation \ref{eq:ppo}
            \STATE $\theta^{old}_{\mathcal{T}_i} \leftarrow \theta_{\mathcal{T}_i}$
        \ENDFOR
        \STATE Freeze module $m_{\mathcal{T}_i}$
        \ENDFOR
         \WHILE{not done}
         \STATE Calculate progress for all tasks using ERP 
         \STATE Choose task $\mathcal{T}_i$ with maximum ERP 
         \STATE Unfreeze module $m_{\mathcal{T}_i}$
         \FOR{$p \in \{1,...,P\}$} 
                 \STATE Sample $\kappa$ episodes by running policy $\pi_{\theta^{old}_{\mathcal{T}_i}}$
                \STATE  $\forall t,$ Compute Advantage estimates $\hat{A}_t$
        \ENDFOR
         \STATE Record average episodic return $R_{m_{\mathcal{T}_i}}(k)$
         \STATE Optimize $L^{\mathcal{T}_i}_{PPO}$ w.r.t. $\theta_{\mathcal{T}_i}$ by Equation \ref{eq:ppo}
         \STATE $\theta^{old}_{\mathcal{T}_i} \leftarrow \theta_{\mathcal{T}_i}$
         \STATE Freeze module $m_{\mathcal{T}_i}$
         \FORALL{$\mathcal{T}_j \in T - \{\mathcal{T}_i\}$}
         \FOR{$p \in \{1,...,P\}$} 
         \STATE Sample $\kappa$ episodes by running policy $\pi_{\theta^{old}_{\mathcal{T}_j}}$
         \ENDFOR
          \STATE Record average episodic return  $R_{m_{\mathcal{T}_j}}(k)$
         \ENDFOR
         \ENDWHILE
    \end{algorithmic}
\end{algorithm}

\subsection{Evaluation Metrics}
\label{sssec:evaluationmetrics}
Designing a reward function is crucial and challenging in deep reinforcement learning, primarily because the reward function may not fully encapsulate all attributes expected from a learning agent. One example is in the Reacher environment, where the need often arises to tune the reward function coefficients or introduce additional parameters. However, this requires significant time and resources to adhere to the predetermined metrics and careful tuning of the parameters. In this sense, evaluating how a straightforward, uninformed reward function performs in terms of metrics meaningful for the task domain at hand is important. Therefore, in this section, in addition to the classical RL metric of episodic return, we also present task domain metrics used to evaluate the collective performance of morphologically different agents. To account for early stopping introduced in \cite{ada_generalization_2022}, we evaluate all tasks after each iteration, save the best policies obtained up to that iteration, and use them in our evaluations. 

\paragraph{Episodic Return}
Maximum episodic return tracks the best expected discounted cumulative reward achieved across all tasks in an iteration. We report the best expected discounted cumulative reward after each training iteration to ensure a fair comparison with baselines. Crucially, the ERP task-switching procedure provides an inherent early-stopping procedure, provided that we allow a fixed number of cumulative training iterations under resource constraints. Since the task is to reach a given point in space with minimum effort, we follow the reward function definition in \cite{towers_gymnasium_2023}
\begin{equation}
\mathbf{r} = -\|p_\text{fingertip} - p_\text{target}\|_2 - \sum \text{a}^2
\label{eq:rewfunc}
\end{equation}
where  $\text{a}$ is the action, $-\sum \text{a}^2$ is the control cost, $p_\text{fingertip}$, and $p_\text{target}$ are the positions of the fingertip and the target, respectively.

\paragraph{Distance To the Goal}
Distance to the goal is the minimum expected L2-norm distance between the final end-effector position and the goal position obtained over all tasks. The Reacher environment only terminates after 50 timesteps, which is equal to the episode length. Hence, we expect the end-effector of the reacher to stay in the immediate vicinity of the goal until episode termination.

\paragraph{Deviation from Shortest Path to the Goal}
At a high level, we expect an efficient learning agent to follow the shortest feasible path to the goal, which is also desirable for many robotic applications. Thus, we introduce a straightness metric that measures the minimum expected deviation from the shortest path to the goal taken by the manipulator's end-effector over all tasks. We define the deviation metric, $\mathcal{D}$ across tasks as

$$
\mathcal{D}=E_\mathcal{T} \left[ \left(\sum_{t=1} L_2(\boldsymbol{x}_\mathcal{T}(t), \boldsymbol{x}_\mathcal{T}(t-1)) \right) - L_2(g, \boldsymbol{x}_\mathcal{T}(0))\right]
$$
where $\mathcal{T}$ is a task, $\boldsymbol{x}_\mathcal{T}(t)$ is the position of the end-effector for task $\mathcal{T}$ at timestep $t$, g is the goal and $\boldsymbol{x}_\mathcal{T}(0)$ is the initial end-effector position. Here, we first compute the length of the spatial trajectory by summing the $L_2$-norm of the distance between the current and the previous end-effector position. Subsequently, we subtract the $L_2$-norm of the distance between the goal position and the initial end-effector position to obtain the deviation from the shortest path to the goal for task $\mathcal{T}$. After each iteration, we calculate the path length traveled by all agents for every task and record the smallest expected deviation from the optimal path to the goal as obtained by the learning algorithm.

\subsection{Implementation Details}

Given the morphological diversity among robots results in different state space dimensions, and our BPNN architecture has a static input layer; we have standardized the state input dimensions for robots with 2-DoF and 3-DoF to match the maximum state space dimension, which of the 4-DoF robot, across all tasks. Hence, we apply zero-padding to the dimensions corresponding to the angular velocity and the angle of each missing link. To encourage skill transfer among networks, we limit the amount of available computational resources to the task networks by setting the number of hidden layers as three and the hidden layer size as two.

For the Reacher tasks, we use PPO \cite{schulman_proximal_2017} as the RL algorithm and adopt the hyperparameter set from \cite{rl-zoo3} for the remaining hyperparameters. After each training iteration, we sample trajectories from each task, excluding the task trained most recently, using the latest BPNN actor and critic parameters. Then, we compute the ERP, maximum episodic return, minimum expected distance to the goal, and minimum expected deviation from the shortest path to the goal for each task. Additionally, we run eight parallel environments per task to reduce simulation time. Each environment runs two episodes, with each episode lasting 50 steps, thereby accumulating a total of 800 timesteps per task.

\section{EXPERIMENTS}
\label{sec:EXPERIMENTS}
In this section, we first present the environment used for benchmarking multi-task learning for morphologically different robots illustrated in Fig.~\ref{fig:robotfigs}. Then, we elaborate on ERP-based task switching and present results obtained with the evaluation metrics introduced in Section \ref{sssec:evaluationmetrics}. 

\subsection{Single Goal Multi-Task Learning Between Morphologically Different Robots}
To demonstrate skill transfer between morphologically different robots, we modified the Reacher-v2 environment, simulated in MuJoCo \cite{todorov_mujoco_2012} using Gymnasium framework \cite{towers_gymnasium_2023}. Our setup consists of three different reacher environments, each having distinct action spaces, specifically featuring 2-DoF, 3-DoF, and 4-DoF robot arms. Although each environment has unique environmental dynamics due to its distinct morphologies, it shares the same reward functions. The reward function defined in Eq. \ref{eq:rewfunc} comprises a control cost, the negative vector norm of the end-effector of the reacher, and a predetermined goal. We consider the baselines RANDOM-BPNN and random Multi-Layer Perceptron (RANDOM-MLP) to evaluate the performance of our ERP-BPNN approach across various metrics. RANDOM-BPNN abides by random task selection while featuring the same underlying BPNN architecture as ERP-BPNN. Similarly, RANDOM-MLP follows the same random task selection procedure as RANDOM-BPNN, but utilizes a separate actor-critic network pair for each task with no lateral connections among different tasks. Accordingly, RANDOM-MLP can be regarded as training a PPO algorithm for each task separately. 

\subsection{Results}
\begin{figure}[htbp]
	\centering
        \subfigure[]{
		\includegraphics[width=0.3\columnwidth]{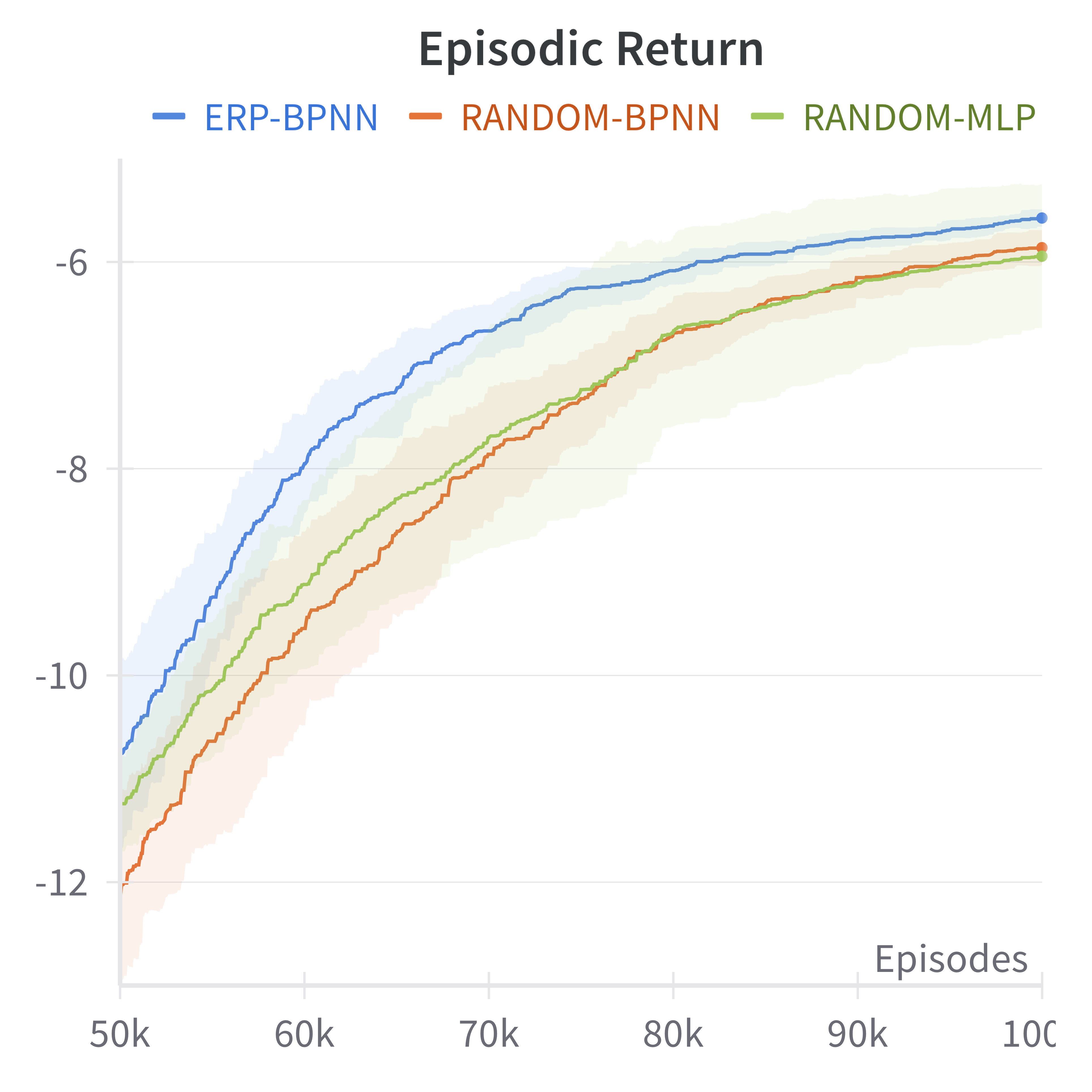}}
  \subfigure[]{
		\includegraphics[width=0.3\columnwidth]{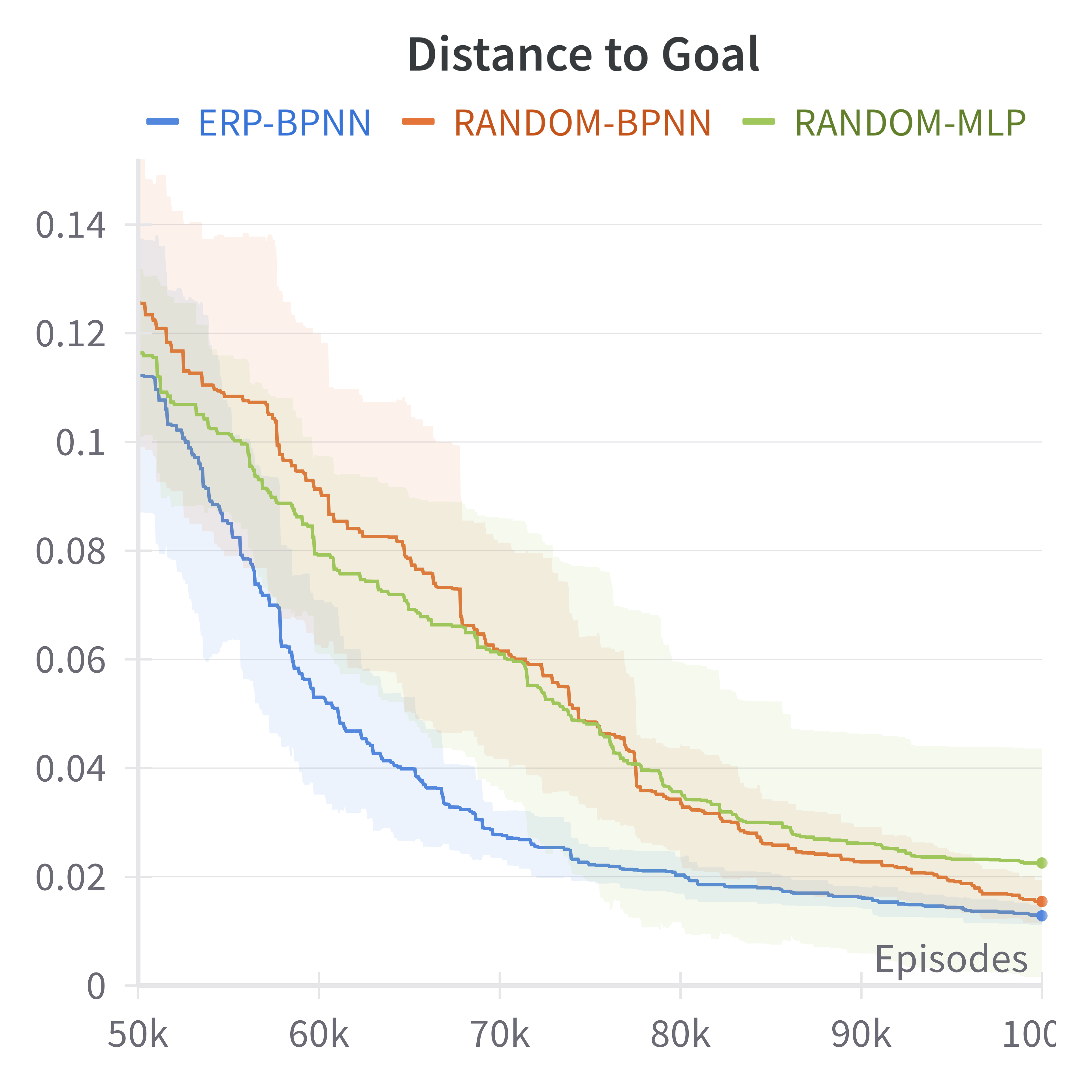}}
  \subfigure[]{
		\includegraphics[width=0.3\columnwidth]{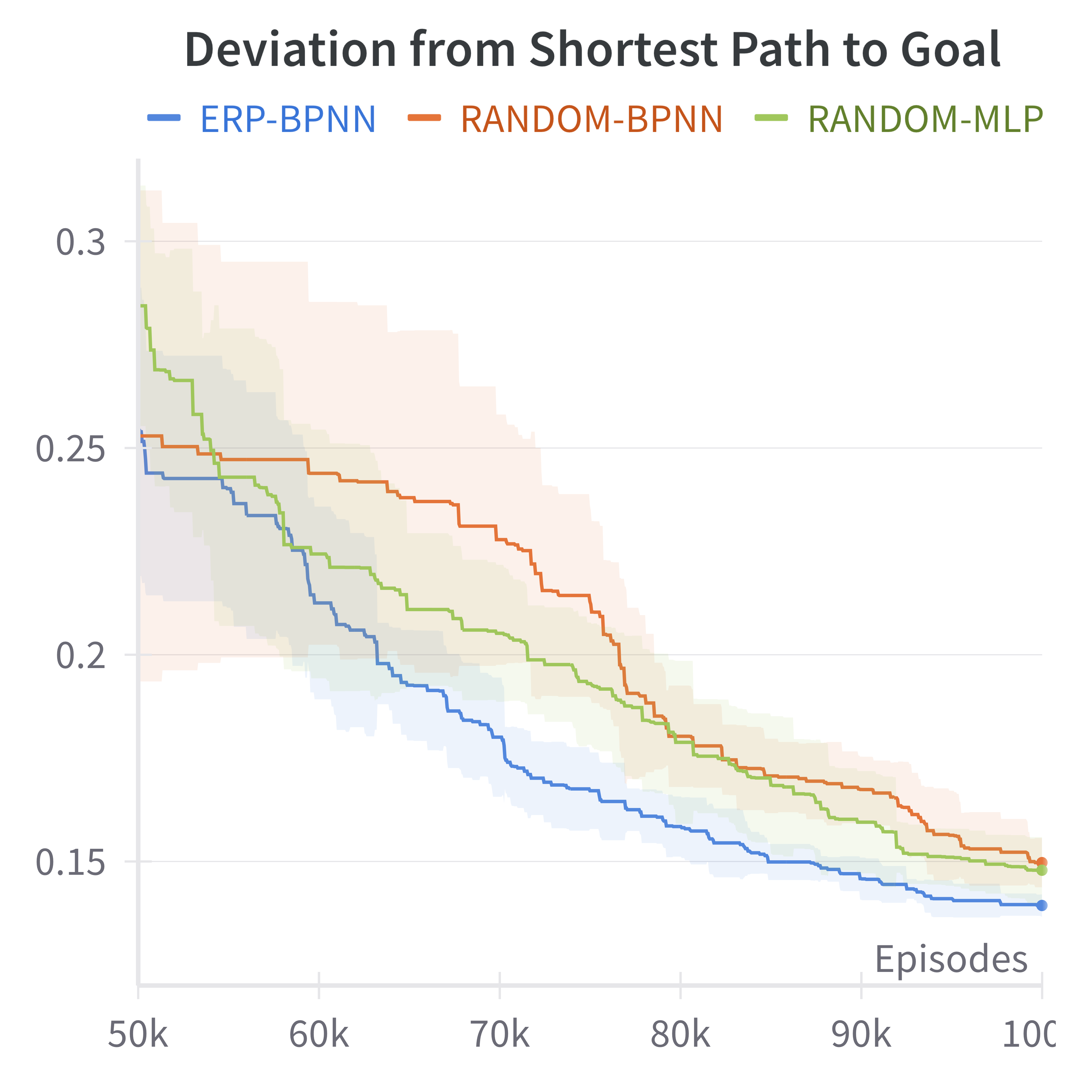}}
	\caption{Performances of the proposed model, ERP-BPNN, and the two baselines of  RANDOM-BPNN and RANDOM-MLP across five random seeds are shown in terms of (a) maximum episodic return, (b) minimum expected final end-effector distance to goal, and (c) minimum expected deviation from the shortest path to the goal.}
	\label{fig:metricplot}
\end{figure}

In this section, we report the results of our method ERP-BPNN compared to the baselines of RANDOM-BPNN and RANDOM-MLP, based on the evaluation metrics presented in Section~\ref{sssec:evaluationmetrics}. Table \ref{table:resultstable} shows each method's mean and standard deviation results obtained during 100K episodes across five random seeds for each metric. The results indicate that ERP-BPNN can obtain superior performance compared to the baselines across all metrics with the lowest standard deviation. More importantly, as can be seen from the Episodic Return plot in Fig. \ref{fig:metricplot} (a),  the proposed model, ERP-BPNN, achieves faster convergence than the baselines. 

\begin{table}[!t]
\caption{Episodic Return, Distance to Goal, and Deviation from Shortest Path to Goal results across 5 random seeds using ERP-BPNN, RANDOM-BPNN, and RANDOM-MLP}
\label{table:resultstable}
\centering
\scalebox{0.77}{
\begin{tabular}{|c||c|c|c|}
\hline
\textit{Episodic Return} & \textbf{ERP-BPNN}  & \textbf{RANDOM-BPNN} & \textbf{RANDOM-MLP} \\
\hline\hline
\text{60,000 episodes}& $\mathbf{-7.95 \pm 0.48}$ & $-9.55 \pm 0.94$ & $-9.12 \pm 0.82$\\\hline
\text{75,000 episodes}& $\mathbf{-6.26 \pm 0.20}$ & $-7.32 \pm 0.45$ & $-7.24 \pm 1.15$\\\hline
\text{80,000 episodes}& $\mathbf{-6.09 \pm 0.13}$ & $-6.72 \pm 0.34$ & $-6.69 \pm 0.91$\\\hline
\text{100,000 episodes}& $\mathbf{-5.58 \pm 0.08}$ & $-5.86 \pm 0.17$ & $-5.95 \pm 0.69$\\\hline
\hline\hline
\textit{Distance to Goal} & \textbf{ERP-BPNN}  & \textbf{RANDOM-BPNN} & \textbf{RANDOM-MLP} \\
\hline\hline
\text{60,000 episodes} & $\mathbf{5.3 \pm 1.79}$&$9.13 \pm 2.86$ &$7.92 \pm 1.82$\\\hline
\text{75,000 episodes} & $\mathbf{2.25 \pm 0.33}$&$4.85 \pm 1.58$ &$4.82 \pm 2.89$\\\hline
\text{80,000 episodes} & $\mathbf{2.03 \pm 0.34}$&$3.42 \pm 0.97$ &$3.57 \pm 2.39$\\\hline
\text{100,000 episodes} & $\mathbf{1.29 \pm 0.18}$&$1.55 \pm 0.4$ &$2.25 \pm 2.1$\\\hline
\hline\hline
\textit{Deviation from Shortest Path} & \textbf{ERP-BPNN}  & \textbf{RANDOM-BPNN} & \textbf{RANDOM-MLP} \\
\hline\hline
\text{60,000 episodes}& $\mathbf{21.25 \pm 2.33}$&$24.39 \pm 4.15$ &$22.44 \pm 3.01$ \\\hline
\text{75,000 episodes}& $\mathbf{16.71 \pm 0.93}$&$21.29 \pm 2.32$ &$19.3 \pm 1.48$ \\\hline
\text{80,000 episodes}& $\mathbf{15.85 \pm 0.73}$&$18.03 \pm 1.23$ &$17.88 \pm 1.97$ \\\hline
\text{100,000 episodes}& $\mathbf{13.94 \pm 0.27}$&$14.97 \pm 0.6$ &$14.79 \pm 0.8$ \\\hline
\hline
\end{tabular}}
\end{table}

Since the combination of BPNN architecture with ERP-based task selection (ERP-BPNN) yields the best results surpassing the random task selection strategy (RANDOM-BPNN), it can be argued that the ERP task selection procedure is essential for successful multi-task learning with positive transfer among tasks. Note that interestingly, the random task selection strategy (RANDOM-BPNN) although inferior to ERP-BPNN, performs better than the RANDOM-MLP baseline in terms of Episodic Return and yields better endpoint accuracy measured as Distance to Goal at the end of the training (Table \ref{table:resultstable}). This indicates the lateral connections among task networks may facilitate limited positive skill transfer even with random task selection. However, this picture is not reflected in the Deviation from the Shortest Path to the Goal measure as random task selection leads to negative transfer for the Deviation from the Shortest Path to the Goal measure (see Fig~\ref{fig:metricplot}(c), episodes $>55$K). In the early stages of learning, the performance of ERP-BPNN and RANDOM-BPNN are better than RANDOM-MLP (episodes $<55K$); however, while the proposed ERP-BPNN enjoys positive skill transfer and thus continues to improve its progress, RANDOM-BPNN suffers from negative interference degrading to a performance worse than RANDOM-MLP in this metric (Fig~\ref{fig:metricplot}(c), episodes $>55$K).

\begin{figure}[htbp]
	\centering
        \subfigure[Typical end-effector paths followed with the policies learned by ERP-BPNN (ours), and the baselines of RANDOM-BPNN and RANDOM-MLP.]{
		\includegraphics[width=\columnwidth]{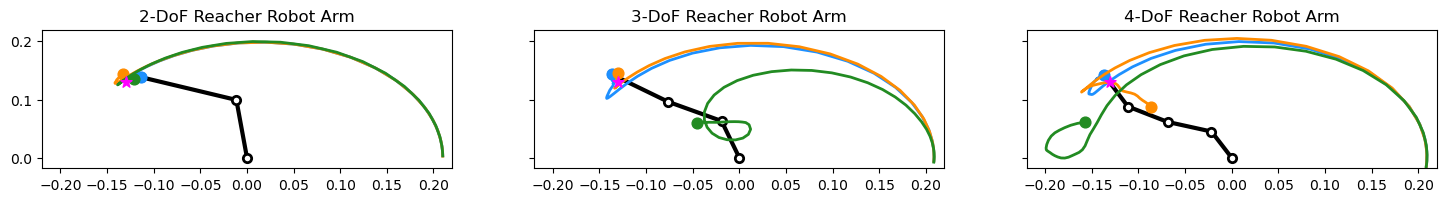}
            }
        \subfigure[Typical end-effector accuracy exhibited with the policies learned by ERP-BPNN (ours), and the baselines of RANDOM-BPNN and RANDOM-MLP.]{
		\includegraphics[width=\columnwidth]{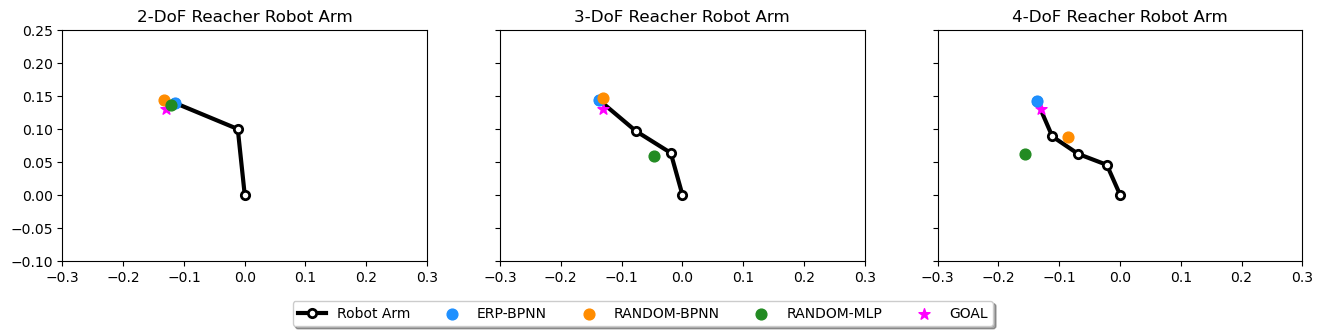}}
	\caption{The policies obtained by our model and the baselines during multi-task learning in terms of straightness (a) and endpoint accuracy (b) are demonstrated (at $1750^{th}$ policy update).  }
	\label{fig:trajgoal}
\end{figure}

In order to get an intuitive understanding of the learned policy performances, we illustrate typical trajectories followed by the end-effector of the Reacher robot in Fig.~\ref{fig:trajgoal}(a) when controlled by policies acquired by the $1750^{th}$ learning iteration, corresponding to approximately 80K episodes, i.e., when approximately 80\% of the training is completed. Observe that at this training stage, the proposed ERP-BPNN has already acquired the skill to reach the target accurately through a less curved path compared to the baselines. In particular, the RANDOM-BPNN diverges from the shortest path to the goal for the 4-DoF robot arm, whereas the RANDOM-MLP diverges from the shortest path to the goal for both 3-DoF and 4-DoF Reacher Robot arms. 

To have an intuitive grasp of the reaching ability obtained on morphologically different robots by the proposed model and the baselines, a typical goal position and the final end-effector positions reached for each robot are shown in Fig.~\ref{fig:trajgoal}(b). Final end-effector positions reached using each method indicate that the agent trained using ERP-BPNN is consistently closest to the goal across all environments compared to the baselines. Likewise, the results for Deviation from Shortest Path to Goal in Fig. \ref{fig:trajgoal}(a) show that the agent trained with ERP-BPNN can reach the goal by taking a shorter path than the baselines in the 4-DoF reacher robot arm environment. Importantly, accurate reaching using a straighter path can be learned significantly faster by our proposed method ERP-BPNN compared to the baselines as evidenced by the distance plots given in Fig. \ref{fig:metricplot}(b).

\begin{figure}[!htbp]
\centering
		\includegraphics[width=0.95\columnwidth]{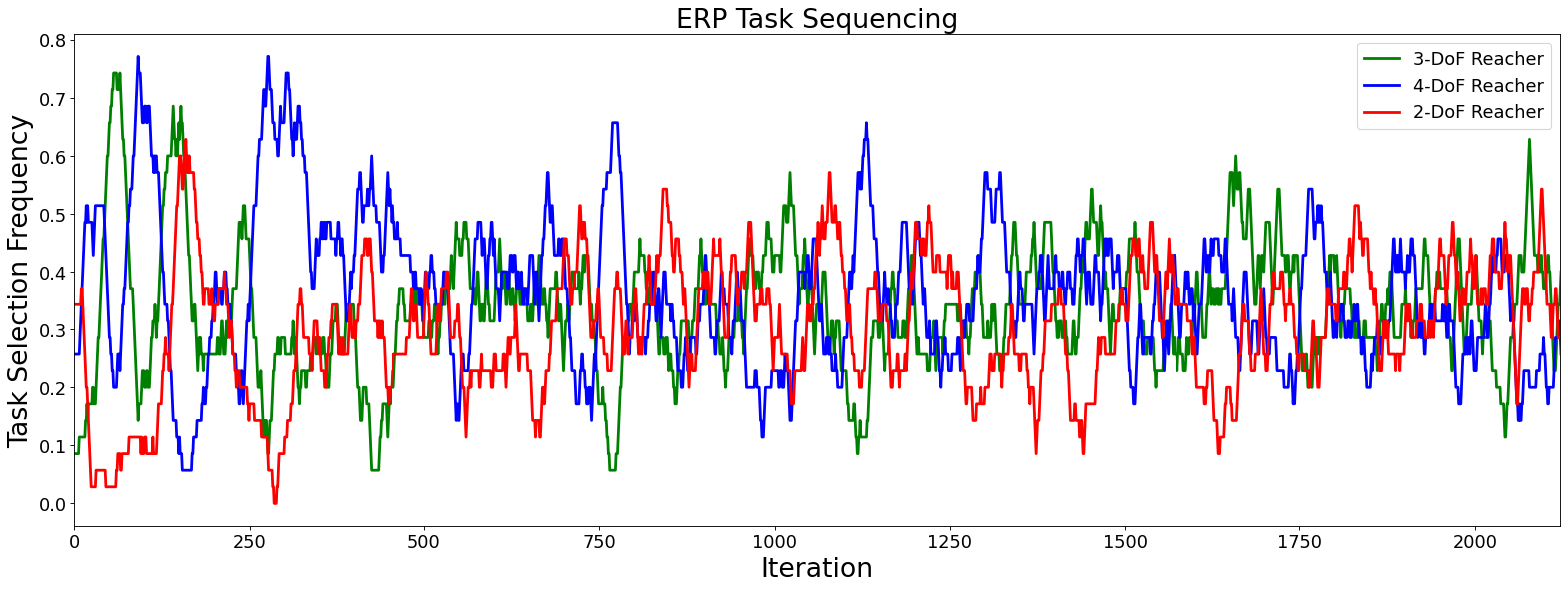}
            
	\caption{Selection frequency plot of task switching by average episodic return progress with an iteration window size of $\nu=35$}
	\label{fig:erptaskseq}
\end{figure}

\subsection{ERP-based Dynamic Task Selection}
To examine the task engagement patterns that emerge with the ERP-based task selection mechanism, we record the selection frequency of each task over a moving learning window, i.e., a fixed number of learning iterations. By using this scheme, the selection frequencies of the tasks are plotted against training iteration counts in Fig.~\ref{fig:erptaskseq}.  As can be seen, although ERP primarily selects the more straightforward 2-DoF Reacher Robot Arm as the one to learn initially (Fig. ~\ref{fig:erptaskseq}, time range: 0-10), it gradually starts selecting more complex tasks with 3-DoF and 4-DoF Reacher Robots, respectively (Fig. ~\ref{fig:erptaskseq} time range: 10-150). Then, we observe that ERP continues selecting the 4-DoF Reacher Robot for skill transfer more frequently (Fig. ~\ref{fig:erptaskseq}, time range: 250-500). This task selection behavior suggests that the 4-DoF Reacher Robot has benefited from the skill transfer of the other tasks and started to learn more rapidly later in the training. Upon further elaboration of the episodic return plots for all tasks in Fig. \ref{fig:metricplot} (a), and standard deviations across iterations in Table \ref{table:resultstable}, ERP-BPNN exhibits a consistent overall improvement across training iterations, achieving a faster convergence with lower standard deviation compared to the baselines. In line with this, faster convergence of ERP-BPNN indicates that consecutive selections of the 4-DoF Reacher task have collectively increased the maximum episodic return. 

\section{CONCLUSION}
\label{sec:CONCLUSION}
This paper introduces Episodic Return Progress with Bidirectional Progressive Neural Networks (ERP-BPNN), a novel multi-task reinforcement learning approach incorporating a human-like interleaved learning mechanism, and shows its application to skill transfer across morphologically different robots. ERP-BPNN comprises a Bidirectional Progressive Neural Network (BPNN) that enables efficient, bidirectional skill transfer and a soft Episodic Return Progress (ERP) mechanism for dynamic task selection. The BPNN architecture is specifically designed to enable skill transfer through bidirectional lateral connections, drawing inspiration from the human brain's adeptness at retaining existing knowledge while acquiring new skills. Episodic return progress-based (ERP) task selection complements BPNN by autonomously selecting the tasks to engage in learning during training, eliminating the need for prior domain knowledge to evaluate difficulty levels of tasks. We demonstrate that soft ERP-based task selection, with BPNN, achieves higher performance and faster convergence than the baselines across all the tested metrics of  Episodic Return, Distance To the Goal, and Deviation from Shortest Path to Goal. 

The insights presented in this work are closely related to the field of human brain-inspired reinforcement learning (RL) research. By implementing an Intrinsic Motivation (IM) signal and leveraging curiosity-driven exploration within the context of multi-task RL, ERP-BPNN framework closely aligns with biological processes observed in the human brain. Specifically, the behavior of dopaminergic neurons, known for their critical role in reward-based learning, mirrors the temporal difference (TD) prediction errors utilized in RL algorithms \cite{odoherty_temporal_2003,schultz_neural_1997,hassabis_neuroscience-inspired_2017}. BPNN provides a mechanism for a shared multi-task RL architecture that prevents task interference during skill transfer by adjusting knowledge transfer from other tasks during training. In addition, the ERP-guided freezing of lateral connections among task modules prevents interference in the previously acquired knowledge. Furthermore, the bidirectional lateral connections augment noise in the task-specific computational flow akin to noisy computations in the brain \cite{khetarpal_towards_2022}, alleviating catastrophic forgetting \cite{ajemian2013} while amplifying plasticity \cite{dohare2021}. Analogous to the modular control architecture in cerebellum \cite{wolpert1998}, as corroborated by Functional Magnetic Resonance Imaging studies \cite{imamizu1997,imamizu2000}, BPNN adopts a modular architecture. This architecture choice is instrumental in decreasing catastrophic forgetting via instantiating a module for each task.

Our plan for future work is to enhance our framework to accommodate lifelong RL scenarios. By integrating additional task modules and further leveraging the BPNN architecture's bidirectional skill transfer capabilities, we aim to enhance our framework's generalizability and learning capacity. Simultaneously, we will refine our soft-ERP task selection procedure to accommodate both previously learned tasks and new tasks. Furthermore, we aim to expand the capabilities of our framework to real-world lifelong learning robotics tasks.



\section*{ACKNOWLEDGMENT}
This research has been supported by JSPS KAKENHI Grant Number JP90542217 and the project JPNP16007 commissioned by the New Energy and Industrial Technology Development Organization (NEDO). In addition, partial support has been given by the INVERSE project (no. 101136067) funded by the European Union.

\bibliographystyle{IEEEtran}  
\bibliography{IEEEabrv,references2}

\end{document}